\documentclass{sig-alternate-05-2015}

\usepackage{subfigure}

\usepackage[latin1]{inputenc}
\inputencoding{utf8}

\newcommand{\norm}[1]{\left\lVert#1\right\rVert}

\DeclareMathOperator*{\argmin}{arg\,min}
\DeclareMathOperator*{\argmax}{arg\,max}

\usepackage[acronym]{glossaries}
\newacronym{ransac}{RANSAC}{Random Sample Consensus}

\begin{document}






%
\conferenceinfo{22nd ACM SIGKDD Conference on Knowledge Discovery and Data Mining}{2016 San Francisco, California USA}

\title{Estimating Phenotypic Traits From \\UAV Based RGB Imagery}

%
%
%
%
%

\numberofauthors{6}
\author{
%
%
\alignauthor
Javier Ribera\\
       \affaddr{Video and Image Processing Laboratory (VIPER)}\\
       \affaddr{Purdue University}\\
       \affaddr{West Lafayette, Indiana USA}
\alignauthor
Fangning He\\
       \affaddr{Digital Photogrammetry Research Group (DPRG)}\\
       \affaddr{Purdue University}\\
       \affaddr{West Lafayette, Indiana USA}
\alignauthor
Yuhao Chen\\
       \affaddr{Video and Image Processing Laboratory (VIPER)} \\
       \affaddr{Purdue University}\\
       \affaddr{West Lafayette, Indiana USA}
\and        
\alignauthor
Ayman F. Habib\\
       \affaddr{Digital Photogrammetry Research Group (DPRG)} \\
       \affaddr{Purdue University}\\
       \affaddr{West Lafayette, Indiana USA}
\alignauthor
Edward J. Delp\\
       \affaddr{Video and Image Processing Laboratory (VIPER)} \\
       \affaddr{Purdue University}\\
       \affaddr{West Lafayette, Indiana USA}
}

\maketitle


\begin{abstract}
In many agricultural applications one wants to  characterize physical properties of plants and use the measurements to predict, for example biomass and environmental influence. This process is known as phenotyping.
Traditional collection of phenotypic information is labor-intensive and time-consuming. 
Use of imagery is becoming popular for phenotyping.
In this paper, we present methods to estimate traits of sorghum plants from RBG cameras on board of an unmanned aerial vehicle (UAV). 
The position and orientation of the  imagery
together with the coordinates of sparse points along
the area of interest are derived through a new
triangulation method. A rectified orthophoto mosaic is
then generated from the  imagery. 
The number of leaves is estimated and a model-based method to analyze the leaf morphology for leaf segmentation is proposed. We present a statistical model to find the location of each individual sorghum plant. 

\end{abstract}

\keywords{image segmentation, image-based phenotyping, image rectification}


\section{Introduction}
In many agricultural applications one wants to  characterize physical properties of plants and use the measurements to predict, for example, biomass and environmental influence. This process is known as phenotyping.
Obtaining high quality phenotypic information is of crucial importance for plant breeders in order to study a crop's performance \cite{white_2012}.
Some phenotypic traits such as leaf area have been shown to be correlated with above-ground biomass \cite{gitelson_2003, golzarian_2011, neilson_2015}.
Collecting phenotypic data is often done manually and in a destructive manner.

Computer-assisted methods and in particular image based methods are becoming popular\cite{li_2014,sankaran_2015}.
There are  commercial image based systems for automated plant phenotyping \cite{lemnatech, granier_2006,fahlgren_2015_2} that are mainly used indoors.
In fact, most image-based phenotyping methods that have been proposed  are for indoor or greenhouse settings,
while real-life crops are mainly grown outdoors \cite{kelly_2015}.
Many different sensor types have been proposed for phenotyping including multi-spectral, hyper-spectral, IR and RGB images.
In this paper we focus on methods that use visible RGB images acquired by a low cost Unmanned Aerial Vehicle (UAV) flying over a field. Our target crop is Sorghum.

In \cite{tsaftaris_2009}, a camera-based growth chamber is described for plant phenotyping using  micropots.
The plant leaves are segmented by selecting a green cluster  in the YIQ colorspace \cite{wu_1992}.
Leaf morphology is determined using morphological processing and  connected components.
The number of leaves per plant and the length of each leaf are estimated.
In \cite{fahlgren_2015_2} the plant \textit{Setaria} is phenotyped in a highly controlled  setting.
The phenotypic traits are estimated from RGB images include plant height, convex hull, and plant area using morphological and watershed methods.
In \cite{pape_2014}, the circular geometry and  overlap between the leaves of rosette plants are used in order to individually segment each leaf.
In \cite{minervini_2014}, rosette plants in a laboratory are automatically segmented and analyzed using active contours and a Gaussian Mixture Model.
Other image based methods are described in several review papers and web sites  \cite{fahlgren_2015, li_2014,lobet_2013}.


Low-cost UAVs have recently emerged as a promising geospatial data acquisition system  for a wide range of applications including phenotyping \cite{tao_advances_2007,chapman_2014, adamsen_1999}.
Recent advances in both automated UAV flying and large-field-of-view (LFOV)  cameras have increased the utilization of UAVs equipped with low-cost LFOV RGB  cameras \cite{habib_improving_2016}.
Due to the inherent spatial distortion caused by LFOV cameras the images must be corrected before phenotypic measurements can be done.
Commercial software  can automate the process of 3D image 3D correction and reconstruction.
However, utilization of such software for  phenotyping remains  a challenging task due to the fact that agriculture fields usually contain poor and repetitive texture. This can severely impact the relative orientation recovery for UAV images. 

This paper presents  methods that use digital images that are acquired from a low-cost UAV using a large field of view RGB camera (e.g., GoPro Hero 3+ Black Edition).
We describe the calibration of the  camera to determine its internal characteristics.
The position and orientation of the acquired imagery together with the coordinates of sparse points along the area of interest are derived through a proposed triangulation technique.
A rectified orthophoto mosaic is then generated by fusing the derived information from the calibration and triangulation information.
Finally, the distortion-free images and orthophoto mosaic are used to derive  phenotypic traits such as leaf area and count and plant locations.

\section{Ortophoto Generation}
\label{sec:orthomethod}
As indicated above, the images from our UAV need to be spatially corrected due to the camera's large field of view.
The three  steps  are described below for our proposed correction methods.
 First, due to severe image distortions caused by the wide-angle lens,  camera calibration is done to estimate the internal characteristics of the   LFOV camera.
Then, Structure from Motion (SfM) \cite{westoby_2012} is used
 to take advantage of prior information of the UAV flight trajectory and for automated image orientation recovery and image sparse point cloud generation.
 Finally, an RGB orthophoto mosaic image of the entire  sorghum field is generated.
 In order to generate this mosaic, the  image exterior orientation parameters (EOPs) are estimated, the camera calibration parameters, and the generated point cloud are used.

\subsection{Camera Calibration}
The  basis of the camera calibration is bundle adjustment \cite{triggs_2000} that includes additional parameters that describe the camera's internal characteristics.
We  demonstrated the capability of using the USGS Simultaneous Multi-frame Analytical Calibration (SMAC) distortion model for both target-based and feature-based calibration of  LFOV cameras \cite{he_target_based_2015}.
The SMAC model is used here for the calibration of the UAV camera.
For the SMAC model, all pixel locations $(x,y)$ must be referenced to the image coordinate system and then translated to the principal point $(x_p,y_p)$. 
$(\bar{x}, \bar{y})$ are the pixel coordinates after correcting to the principal point using Equation \ref{eq:fang1}.

\begin{equation}
  \centering
  \begin{split}
    \bar{x} = x - x_p
    \\
    \bar{y} = y - y_p
  \end{split}
  \label{eq:fang1}
\end{equation}

For the distortion parameters, both radial and decentering lens distortions are considered.
Radial lens distortion is caused by the large off-axial angle and lens manufacturing flaws and are located along a radial direction from the principal point.
Radial lens distortion is more significant in LFOV cameras.
According to the SMAC model, the correction for the radial lens distortion $(\Delta x_{RLD}, \Delta y_{RLD})$ can be expressed as in Equation \ref{eq:fang2} using the coefficients $(K_0,K_1,K_2,K_3)$.

\begin{equation}
  \small
  \begin{split}
    \Delta x_{RLD} = \bar{x} \left[ K_0 + K_1(r^2-R_0^2) + K_2(r^4-R_0^4) + K_3(r^6-R_0^6) \right]
    \\
    \Delta y_{RLD} = \bar{y} \left[ K_0 + K_1(r^2-R_0^2) + K_2(r^4-R_0^4) + K_3(r^6-R_0^6) \right]
  \end{split}
  \label{eq:fang2}
\end{equation}
where $r=\sqrt{x^2+y^2}$, $K_0, K_1, K_2$, and $K_3$ are the radial lens distortion parameters, $R_0$ is a camera-specific constant.
In this paper,  $R_0$  is defined as 0 mm.
Decentering lens distortion is caused by misalignment of the elements of the lens system along the camera's optical axis.
The decentering lens distortion has radial and tangential components.
To evaluate the decentering lens distortion $(\Delta x_{DLD}, \Delta y_{DLD})$ for the measured points, one can use the coefficients $(P_1,P_2)$ as shown in Equation \ref{eq:fang3}.

\begin{equation}
  \begin{split}
    \Delta x_{DLD} = P_1(r^2+2\bar{x}^2) + 2P_2\bar{x}\bar{y}
    \\
    \Delta y_{DLD} = 2P_1\bar{x}\bar{y} + P_2(r^2+2\bar{y}^2)
  \end{split}
  \label{eq:fang3}
\end{equation}

where $P_1$ and $P_2$ are the decentering lens distortion parameters.


\subsection{Structure From Motion}
\label{sec:SfM}
The SfM \cite{westoby_2012} automates the image EOPs recovery as well as the generation of the sparse image-based point cloud.
The  SfM approach we used was  described by He and Habib in \cite{he_linear_2014}.
It is based on a 3-step strategy.
 In the first step, the relative orientation parameters (ROPs) of all possible image stereo pairs are estimated.
  The estimation of ROPs requires the identification of conjugate points in the set of available images.
  Therefore, we use the Scale-Invariant Feature Transform (SIFT)\cite{lowe_distinctive_2004} on the stereo-images.
  Using the Euclidean distances between the SIFT descriptors, potential matches  are identified \cite{lowe_distinctive_2004}.
  Then, an  approach which uses prior information of the UAV flight trajectory is used for the estimation of the ROPs.
 We assume that the UAV  is moving at constant  altitude while operating a nadir-looking camera.
  This is consistent with our UAV operations using onboard gimbals.
  Therefore a minimum of two point correspondences are required for the estimation of the ROPs.
   \gls{ransac} \cite{fischler_1987} is used  to remove potential outliers among the initial matches.
   \gls{ransac} is used to find the point-to-epipolar distance of each conjugate point pair using parameters that are derived from different randomly selected point samples.
Then an estimate of the ROPs using an iterative procedure is done.
   This approach is different from the conventional non-linear least-square adjustment in that it is based on a direct linearization of the co-planarity model \cite{mikhail_introduction_2001}.
   The matching  outliers are then removed using a normalization process that imposes constraints on the xy-parallax  according to epipolar geometry.
   
   Once the ROPs of all possible stereo-pairs are estimated, the image EOPs are initialized.
   This initialization starts with defining a local coordinate frame using an image triplet that satisfies both a large number of corresponding points and a  compatibility configuration \cite{he_automatic_2014}.
   The remaining images are then sequentially augmented into the final image block or trajectory.
   In order to reduce the effects of error propagation, at each step of the image  augmentation, only the image that exhibits the highest compatibility with the previously referenced imagery is selected.
   Finally, this image is aligned relative to the pre-defined local coordinate system.
   
\begin{figure}[h]
\centering
\includegraphics[width=8.5cm]{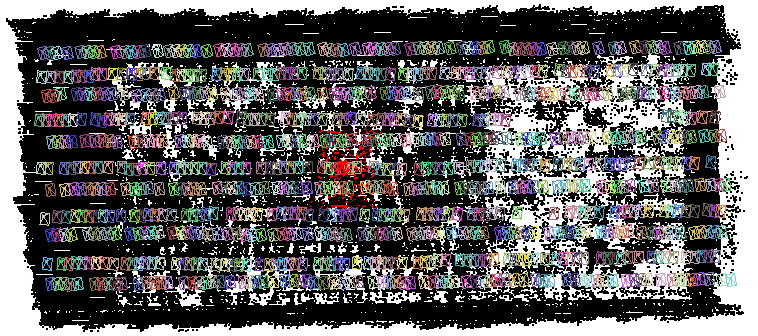}
\caption{Reconstructed sparse point cloud, positions and orientations.}
\label{fig:fang_cloud}
\end{figure}

\begin{figure}[b]
\centering
\includegraphics[width=8.5cm]{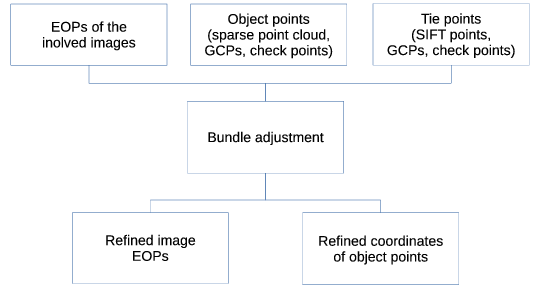}
\caption{The proposed global bundle adjustment process to transform the image EOPs and sparse point cloud coordinates to the mapping frame.}
\label{fig:fang_diagram2}
\end{figure}

   The  EOPs from the initial recovery are only relative to an arbitrarily defined local reference frame.
We transform the estimated  EOPs to an absolute orientation and the  sparse point cloud to a mapping reference frame.
An example is shown in Figure \ref{fig:fang_cloud}.
   The coordinates of  Ground Control Points (GCPs) are derived through a  prior GPS survey of the sorghum field.
   These GCPs are used to establish the absolute orientation.
   After the estimation of the absolute orientation parameters, a global bundle adjustment is done.
   This bundle adjustment includes SIFT-based tie points, manually measured image coordinates of the GCPs, and the corresponding GPS-coordinates of these GCPs.
   This yields the image EOPs and the sparse point-cloud coordinates relative to the GCPs' reference frames.
   Figure \ref{fig:fang_diagram2} illustrates the inputs and outputs of the global bundle adjustment process.

\subsection{RGB Orthophoto Mosaic}
\label{sec:rgb-based-ortophoto}
\begin{figure}[h]
\centering
\includegraphics[width=8.3cm]{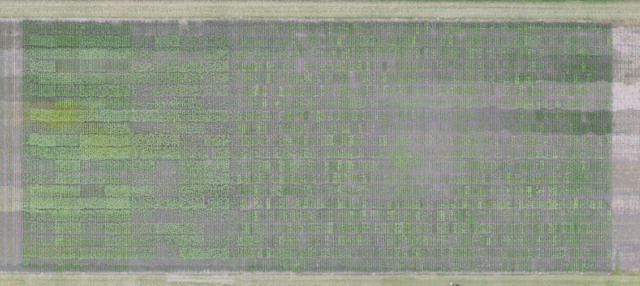}
\caption{RGB orthophoto of the sorghum field.}
\label{fig:fang_field}
\end{figure}

Finally, in order to generate the RGB orthophoto, a Digital Elevation Model (DEM) is interpolated from the sparse point cloud, which has been transformed relative to the mapping frame.
Then, the DEM, the bundle adjustment-based image EOPs, and the camera calibration parameters are used to produce an RGB orthophoto mosaic of the entire  field.
An example of the  ortophoto is shown in Figure \ref{fig:fang_field}.

The conceptual basis for generating an RGB orthophoto mosaic is an object-to-image backward and forward projection process using  collinearity equations \cite{mikhail_introduction_2001}.
It is important to note that, although each pixel from the orthophoto can be visible in multiple UAV images, only the closest one in the mapping frame is utilized for the RGB mosaic.

\section{Image Based Phenotyping}


\subsection{Leaf Counting}
\label{sec:color-based}
In this section, we present a method for leaf counting without individually segmenting each leaf.

First, the uncorrected image or orthomosaic is converted from RGB to HSV color space \cite{agoston_2005}.
From the image in the HSV color space, a segmentation mask $Y$ is generated.
Each pixel $Y_m$ in this mask is obtained as:
 \begin{equation}
   Y_m = \left\{
     \begin{array}{ll}
       1 & \text{ if  } \tau_1 \leq H_m \leq \tau_2 \text{ and }( \tau_3 \leq S_m \text{ or }  \tau_4 \leq V_m )  \\
       0 & \text{otherwise}
     \end{array}
   \right.
\end{equation}
where $H_m$, $S_m$, and $V_m$ are the hue, saturation, and value of the pixel $m$.
The  thresholds $\tau_1, \tau_2, \tau_3,$ and $\tau_4$ are determined experimentally.
$\tau_1$ and $\tau_2$ select the characteristic green color of the leaves.
$\tau_3$ and $\tau_4$ prevent misclassifying some soil pixels as leaves.
Then, the number of pixels classified as sorghum leaves is determined as
 \begin{equation}
   \alpha = \sum_{m=0}^{M-1} Y_m
\end{equation}
where $M$ is the number of pixels in the image.
This pixelwise segmentation makes use of the strong color difference between the sorghum leaves, which are generally green or yellow, and the soil and panicles, which are usually brown.
An example of the segmentation result is shown in Figure \ref{fig:combined_segmentation}.

\begingroup
\makeatletter
\renewcommand{\p@subfigure}{}
\begin{figure}[b]
  \centering
  \subfigure[]
  {
          \includegraphics[width=4cm]{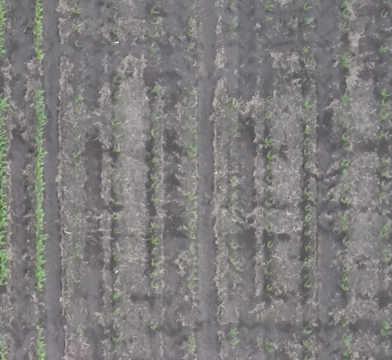}
          \label{fig:combined_segmentation1}
  }
  \hspace{-0.3cm}
  \subfigure[]
  {
          \includegraphics[width=4cm]{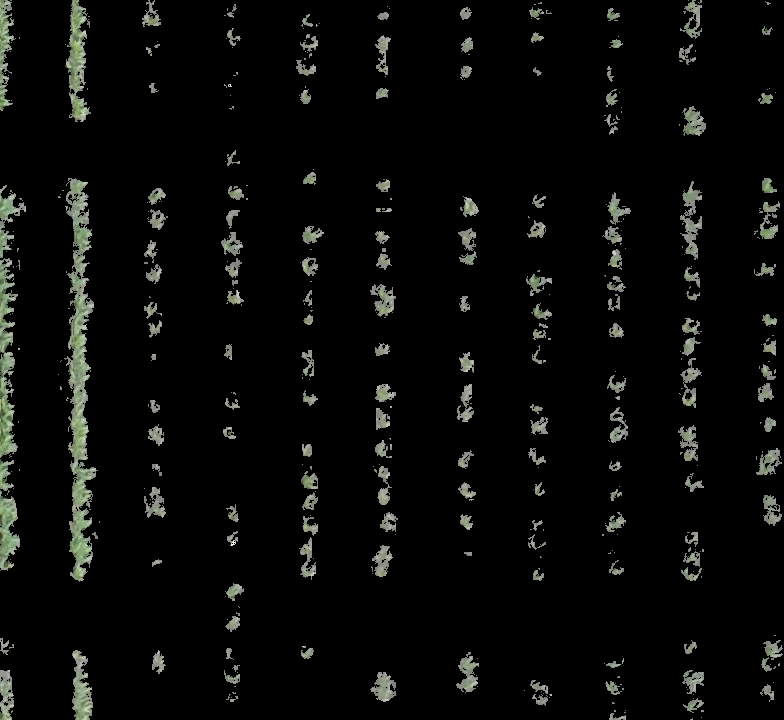}
          \label{fig:combined_segmentation2}
  }
  \hspace{-0.3cm}
  \caption{
    \ref{fig:combined_segmentation1} Section of a orthorectified mosaic  at an altitude of 15 m.
    \ref{fig:combined_segmentation2}  Segmentation result.
  }
\label{fig:combined_segmentation}
\end{figure}
\endgroup

Now, we want to estimate the number of leaves, denoted as $\lambda$, from $\alpha$.
In order to do this, we assume that number of leaves and the number segmented pixels are linearly related as $\lambda = \frac{\alpha}{\rho}$.
This assumes that all leaves have approximately the same area.
The number of pixels per leaf is $\rho$.

In order to calibrate $\rho$, a small region of the image is selected.
The number of leaves in this region is manually counted and denoted as $\lambda_0$.
The number of pixels classified as sorghum is denoted as $\alpha_0$.
Finally, $\rho$ is estimated by $\rho = \frac{\alpha_0}{\lambda_0}$, and the final leaf count can be obtained as $\lambda = \frac{\alpha}{\rho}$.
We experimentally determined that the relationship between the number of leaves and the number of  sorghum pixels  is approximately linear at a given growth stage.
This method requires that all leaves are at the same distance from the camera.
This condition is fulfilled when using the orthorectified mosaic.
Also, only leaves that are visible can be counted.

\subsection{Model Based Leaf Segmentation}
\label{sec:leaf_segmentation}

Plant leaves have similar morphology but different traits such as leaf length, width and green color hue.
From the segmentation of each individual leaf, these traits can be estimated.
In this section, we present a shape-based approach to leaf segmentation.


\begingroup
\makeatletter
\renewcommand{\p@subfigure}{}
\begin{figure}[h!]
  \centering
  \subfigure[]
  {
          \includegraphics[width=2cm]{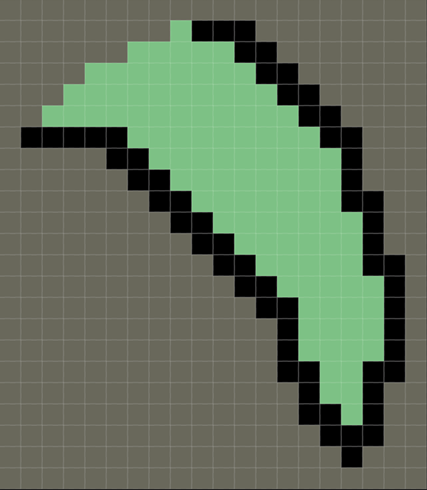}
          \label{fig:leaf_nude}
  }
  \hspace{-0.3cm}
  \subfigure[]
  {
          \includegraphics[width=2cm]{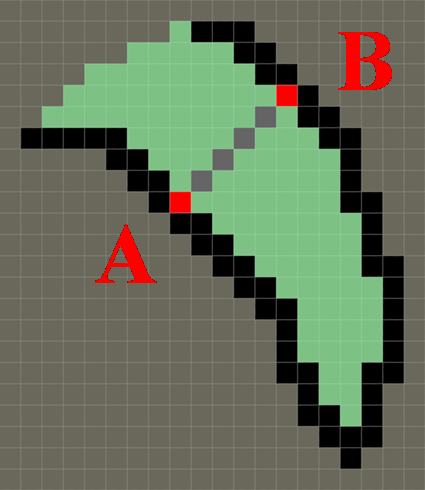}
          \label{fig:leaf_AB}
  }
  \hspace{-0.3cm}
  \subfigure[]
  {
          \includegraphics[width=2cm]{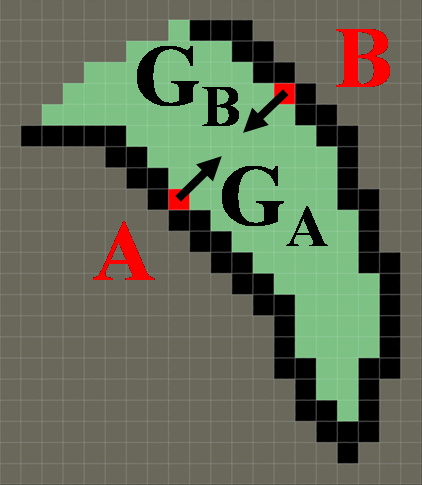}
          \label{fig:leaf_slice}
  }
  \hspace{-0.3cm}
  \subfigure[]
  {
          \includegraphics[width=2cm]{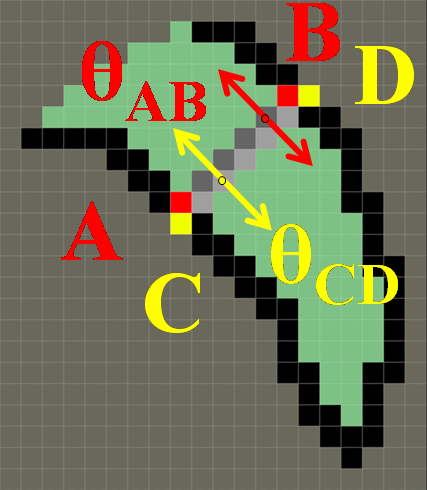}
          \label{fig:leaf_slice_gradient}
  }
  \hspace{-0.3cm}
  \caption{
    \ref{fig:leaf_nude} A synthetic leaf.
    \ref{fig:leaf_AB} A leaf slice connecting two pixels, $A$ and $B$, at opposite edges of the leaf.
    \ref{fig:leaf_slice} $G_A$ and $G_B$ are the gradient angles of pixels $A$ and $B$.
    \ref{fig:leaf_slice_gradient} $\theta_{AB}$ and $\theta_{CD}$ are the angles the of slices $S_{AB}$ and $S_{CD}$.
  }
  \label{fig:leaf}
\end{figure}
\endgroup

$A$ and $B$ are two pixels located at two opposite edges of a leaf.
A pixel-wide line connecting $A$ and $B$ is defined as the leaf slice $S_{AB}$.
Figure \ref{fig:leaf} depicts these definitions with a synthetic leaf.
The gradient angles at $A$ and $B$ are $G_A$ and $G_B$, respectively.
The angle $0\textdegree$ is defined in the direction of the $x$ axis.
$G_A$ and $G_B$ are expected to be opposite to each other with some bias $T_a$, i.e,

\begin{equation}
  |G_A - G_B + \pi| \mod 2\pi < T_a
  \label{eq:leaf_edges}
\end{equation}

Leaf slices are obtained by using the Stroke Width Transform \cite{epshtein_2010}.
The slice angle $\theta_{AB}$ of $S_{AB}$ is defined as the normal of the slice (in radians) as

\begin{equation}
  \theta_{AB} = \frac{G_A + G_B}{2} \mod \pi
\end{equation}

In order to reconstruct leaves from leaf slices, adjacent leaf slices, $S_{AB}$ and $S_{CD}$, are compared.
If their angles $\theta_{AB}$ and $\theta_{CD}$  differ less than a constant  $T_b$, i.e, 

\begin{equation}
  |\theta_{AB} - \theta_{BC}| < T_b
  \label{eq:leaf_slice_merging}
\end{equation}

then slices $S_{AB}$ and $S_{CD}$ are merged.

Plants with high leaf density can cause leaves overlap with each other.
In the case of leaf overlap, there may be a discontinuity between leaf slices as shown in Figure \ref{fig:search_segment}.

\begingroup
\makeatletter
\renewcommand{\p@subfigure}{}
\begin{figure}[h!]
  \centering
  \subfigure[]
  {
          \includegraphics[width=2.6cm]{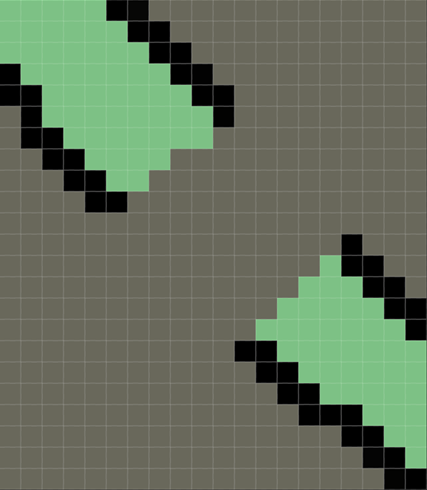}
          \label{fig:search_segment_nude}
  }
  \hspace{-0.2cm}
  \subfigure[]
  {
          \includegraphics[width=2.6cm]{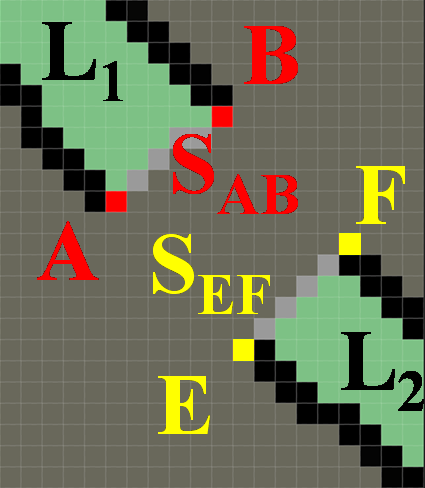}
          \label{fig:search_segment_slices}
  }
  \hspace{-0.2cm}
  \subfigure[]
  {
          \includegraphics[width=2.6cm]{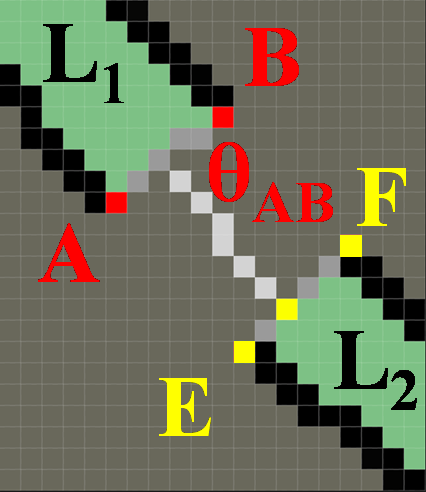}
          \label{fig:search_segment_search}
  }
  \hspace{-0.2cm}
  \caption{
    \ref{fig:search_segment_nude} A single leaf with a discontinuity that may be due to a leaf crossing.
    \ref{fig:search_segment_slices} Two leaf segments $L_1$ and $L_2$ with almost parallel facing slices $S_{AB}$ and $S_{EF}$.
    \ref{fig:search_segment_search} From leaf slice $S_{AB}$, we search in the direction of $\theta_{AB}$ until we find the slice $S_{EF}$ of the opposite leaf segment $L_2$.
  }
  \label{fig:search_segment}
\end{figure}
\endgroup

In this case, the two leaf segments will be merged by the following search method.
Let a leaf be split into two leaf segments $L_1$ and $L_2$, separated by a discontinuity, as shown in Figure \ref{fig:search_segment}.
Let $S_{AB}$ be the leaf slice with angle $\theta_{AB}$ at the end of $L_1$.
Let $S_{EF}$ be the leaf slice with angle $\theta_{EF}$ at the end of $L_2$.
Let $X_{AB}$ to be a vector contains all the pixels in $S_{AB}$.
From pixel $x_i$ in $X_{AB}$, we search in the direction of $\theta_{AB}$.
If a leaf slice $S_{EF}$ from another leaf $L_2$ is found and the difference between the two angles is less than a constant  $T_c$, i.e,
\vspace{-5pt}
\begin{equation}
  |\theta_{AB} - \theta_{EF}| < T_c     \text{, }
\end{equation}
then leaves segments $L_1$ and $L_2$ are considered the same leaf. 

Note that the   thresholds $T_a, T_b, T_c$ are determined experimentally.
This method addresses leaf discontinuity due to leaf crossing, but requires that leaf edges are clearly distinguishable.
Overlap of parallel leaves may lead to undersegmentation.


\subsection{Plant Location Model}
\label{sec:plant_location}
In this section we present a statistical model for segmentation of the plant and determining its location.
The location of each plant is defined as the pixel coordinates where the stalk intersects the ground plane and
can be used to automatically obtain the inter-row and intra-row spacing.
A row of plants is defined as all the plants that are aligned together.
Inter-row spacing is defined as the distance between rows.
Intra-row spacing is defined as the distance between plants within the same row.

The number of plants in an image is denoted as the constant $P$ and assumed to be known a priori.
The positions of the plants are modeled as a random vector $X$, i.e,
\begin{equation}
  X = [X_0, X_1, ..., X_{P-1}]
\end{equation}
where $X_p$, $p=0,...,P-1$, contains the $(i,j)$ coordinates of the $p$-th plant:
\begin{equation}
  X_p = \begin{bmatrix}
           X_{p,i} \\
           X_{p,j} \\
        \end{bmatrix}
\end{equation}

Our goal is to estimate $X$ from $Y$, where is $Y$ is the color based segmentation mask (Section \ref{sec:color-based}).


The 2D coordinates of the pixel $m$ are denoted as $K(m)$. A vector $Z$ is constructed as 
\begin{equation}
  Z = [Z_0, Z_1, ..., Z_{N-1}]
\end{equation}
where each element $Z_n = K(n)$, $n=0,...,N-1$, is included if $Y_n=1$.
$N$ is the number of pixels classified as leaf pixels.
Notice that $N\leq M$.

\begin{figure}[h]
\centering
\centerline{\includegraphics[width=3cm]{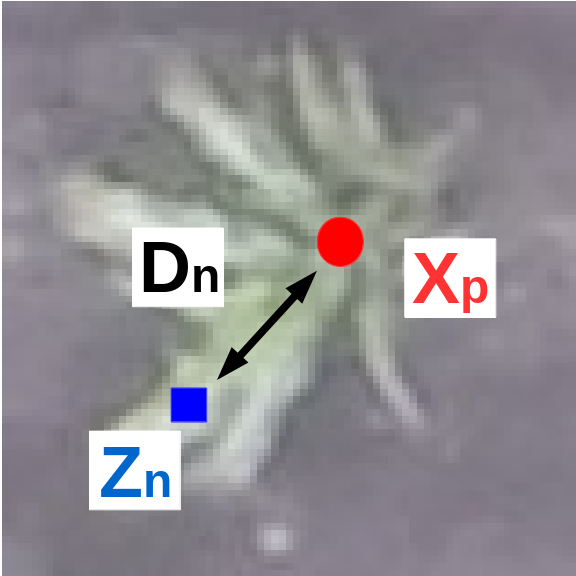}}
\caption{
A single sorghum plant.
The distance from pixel $n$ (with coordinates $Z_n$) to $X_p$ (the coordinates of Sorghum plant $p$) is $D_n$, obtained by Equation \ref{eq:dist}.}
\label{fig:forward}
\end{figure}

The plant $p$ that is closest to the pixel $n$ is denoted as $C(n)$.
The Euclidean distance from the pixel $n$ to the plant $C(n)$ is denoted as $D_n$ and is computed as in Equation \ref{eq:dist}.
Figure \ref{fig:forward} depicts the location $X_p$ of one sorghum plant, and the distance $D_n$ to a leaf pixel $n$.
\begin{equation}
\begin{split}
  D_n & = \norm{K(n)-K(C(n))}_2 \\
      & = \argmin_{p=0,1,...,P-1} \norm{K(n)-X_p}_2
\end{split}
  \label{eq:dist}
\end{equation}

 $D_n$ is modeled as a random variable with exponential conditional probability density with mean and standard deviation $\sigma$.
Therefore the probability density function for a leaf pixel $n$  at distance $D_n = d_n$ from $C(n)$ is
\begin{equation}
  p_{D_n}(d_n) = \frac{1}{\sigma} e^{-\frac{d_n}{\sigma}}
\end{equation}

$\sigma$ can be interpreted as the average radius of a plant.

Then, the conditional distribution of a single point $Z_n$ only depends on its closest plant:
\begin{equation}
\begin{split}
  p_{Z_n | X}(z_n | X) & = p_{Z_n | X_{C(n)}}(z_n | X_{C(n)}) \\
                       & = \frac{1}{\sigma} e^{-\frac{d_n}{\sigma}}
\end{split}
\end{equation}

From our assumptions above we have  conditional independence and the joint conditional density of $Z$ is
\begin{equation}
\begin{split}
  p_{Z | X}(z | X) & = \prod_{n=1}^{N} p_{Z_n | X}(z_n | X)  \\
                   & = \frac{1}{\sigma^N} e^{-\frac{1}{\sigma} \sum_{n=1}^N d_n}
\end{split}
\label{eq:cond}
\end{equation}

This model assumes that the leaf distribution does not have any direction preference, i.e, the leaves grow uniformly in all directions.
In some situations, however, the plant is tilted, and the stalk is not completely at the center of the plant.

As we can see in Figure \ref{fig:combined_segmentation}, since we are using an orthorectified mosaic, the crop field follows a  structure.
The plants in the image are very much aligned in rows as they are in the field.
We make use of this information to introduce a prior distribution for $X$.

\begin{figure}[htb]
  \centering
  \centerline{\includegraphics[width=3cm]{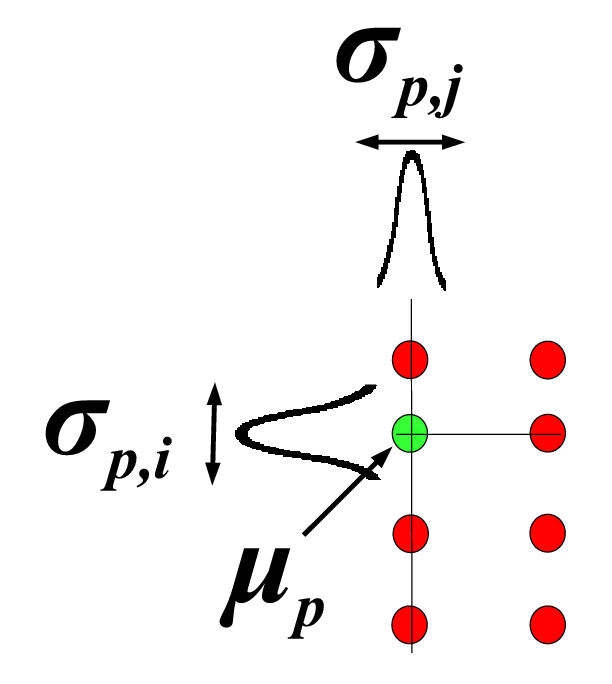}}
  \caption{ Vertical and horizontal alignments of the plants in the field when viewed vertically. The green dot is the plant whose prior position we are developing.}
  \label{fig:prior}
\end{figure}

The conditional probability density of the position of one plant $X_p$ given the remaining plants $X_0, ..., X_{p-1}, X_{p+1}, ..., X_{P-1}$ is assumed normal

\begin{equation}
  p_{X_p | X_{q \neq p}}(x_p | X_{q \neq p}) = \frac{1}{2\pi |R_p|^{1/2}}
                                               \exp{ \left( -\frac{1}{2} \norm{x_p - \mu_p}_{R_p^{-1}}^2 \right) }
\end{equation}

where $\mu_p$ are the coordinates of the vertical and horizontal plant lines where $X_p$ is a member, and

\begin{equation}
  R_p = \begin{bmatrix}
         \sigma_{p,i}^2 & 0 \\
         0 & \sigma_{p,j}^2 \\
       \end{bmatrix}
\end{equation}

is the covariance matrix of the positions of the plants that are aligned with the plant $p$, either vertically or horizontally.
$\sigma_{p,i}^2$ and $\sigma_{p,j}^2$ are the vertical and horizontal standard deviations of $X_p$ (see Figure \ref{fig:prior}).
$\sigma_{p,j}^2$ is typically very low because of the alignment of the planter at planting time.
$R_p$ is a diagonal matrix when the field rows are aligned with the image axis in the orthorectified image.

From Equation \ref{eq:cond}, we can obtain the MAP estimate of $X$ as 
\begin{equation}
  \begin{split}
    \hat{X}(Z) & = \argmax_{x} p_{X|Z}(x|Z)  = ...  \\
                     & = \argmin_{x}  \left( -\ln p_{Z|X}(Z|x) - \ln p_{X}(x) \right) \\
                     & = \argmin_{x}  \left( \frac{1}{\sigma} \sum_{n=1}^{N}d_n - \ln p_{X}(x)  \right)
  \end{split}
\end{equation}

Obtaining a closed form for $p_{X}(x)$ involves dependencies between pant positions because the plant positions are not mutually independent.
We iteratively obtain the MAP estimate of each plant position $X_p$ separately:
\begin{equation}
  \begin{split}
    \hat{X}_{p}(Z) & = \argmax_{x_p} p_{X_p|Z, X_{q \neq p}}(x|Z, X_{q \neq p})  = ...  \\
                   & = \argmin_{x_p}  \left( -\ln p_{Z|X}(Z|x) - \ln p_{X_p | X_{q \neq p}} (x_p | X_{q \neq p})  \right) \\
                   & = \argmin_{x_p}  \left( \frac{1}{\sigma} \sum_{n=1}^{N}d_n + \frac{1}{2}\norm{x_p - \mu_p}_{R_p^{-1}}^2 \right)
  \end{split}
  \label{fig:cost_function}
\end{equation}


For the special case in which the prior term is not used, the estimate $\hat{X}_{p}(Z)$ in Equation \ref{fig:cost_function} is reduced to 
\begin{equation}
  \hat{X}_{p}(Z) = \argmin_{x_p} \sum_{n=1}^{N} d_n
  \label{eq:k-means-cost-function}
\end{equation}

This corresponds to the cost function of the k-means clustering technique.
In this case, $\hat{X}_{p}(Z)$, has a closed form solution, shown in Equation \ref{eq:k-means-cost-function}, which is the average of the points in the cluster formed by  plant $p$.

\begin{equation}
  \hat{X}_{p}(Z) = \frac{\sum_{n=1}^{N} h(x_p | Z_n) Z_n}{\sum_{n=1}^{N} h(x_p | Z_n)} 
  \label{eq:k-means-closed-form}
\end{equation}

where 
 \begin{equation}
   h(x_p | Z_n) = \left\{
                     \begin{array}{ll}
                       1 & \text{if } p = C(n)\\
                       0 & \text{otherwise}
                     \end{array}
                   \right.
\end{equation} 
is the membership function that indicates whether the pixel $n$ corresponds to plant $x_p$ or not.

Another special case occurs when the prior distribution about the intra-row spacing prior is not used. When $\sigma_{p,i} \to \infty$, Equation \ref{fig:cost_function} becomes

\begin{equation}
  \lim_{\sigma_{p,i} \to \infty} \hat{X}_{p}(Z) =
  \argmin_{x_p}  \left( \frac{1}{\sigma} \sum_{n=1}^{N}d_n + \frac{1}{2}\left( \frac{x_{p,j} - \mu_{p,j}}{\sigma_{p,j}} \right)^2 \right)
  \label{eq:no-inter-row-prior}
\end{equation}


Since  $\sigma$ is not known, we estimate it after every Iterative Coordinate Descent (ICD) \cite{wright_2015} iteration using the Maximum Likelihood estimator,
that can be obtained in closed form:

\begin{equation}
  \begin{split}
    \hat{\sigma}(Z,X) & = \argmax_{\sigma} p_{Z|X, \sigma}(Z|X, \sigma)  = ...  \\
                      & = \frac{1}{N} \sum_{n=0}^{N-1} d_n
  \end{split} 
\end{equation}


\section{Experimental Results}

The experimental  dataset was acquired by a DJI Phantom 2 UAV equipped with a GoPro Hero 3+ Black Edition camera
at the Agronomy Center for Research and Education (ACRE) at Purdue University.
The UAV was at an altitude of approximately 15 meters and at a speed of roughly 8 m/s.
The overlap and side lap percentages for the images are approximately 60\%.
The ground sampling distance for the images is roughly 1.5cm.
The  camera is mounted on a gimbal to ensure that images are acquired with the camera's optical axis pointing in the nadir direction, which is consistent with our assumptions for the triangulation approach we described in Section \ref{sec:SfM}.
To generate the orthophoto mosaic, 18 Ground Control Points (GCPs) and 10 check points are established.
Both GCPs and check points were surveyed with a GPS to an approximate accuracy of $\pm 2$ cm.
The root-mean-square error (RMSE) for the check points is evaluated after the  bundle adjustment as described in Section \ref{sec:rgb-based-ortophoto}.
The outcome from the bundle adjustment is finally used to produce an orthophoto mosaic image.
We collected on June 24, 2015 489 images with  RMSE for the X, Y, and Z coordinates being $0.04$ m, $0.03$ m, and $0.05$ m, respectively.
Figure \ref{fig:fang_field} shows the orthophoto mosaic image.
We also collected on July 15, 2015  497 images with  RMSE for the X, Y, and Z coordinates being $0.03$ m, $0.03$ m, and $0.04$ m, respectively.
Figure 11\ref{fig:result_plot_20150715} shows the orthophoto mosaic image.
Note that the height and nadir-looking 
assumptions are only required for stereo-pairs that may 
be captured along the same flight line or in two different flight lines.
Experimental results involving captured datasets by fixed-wing and quad copters have shown that the proposed procedure is quite robust to deviations from such assumptions.

%


The color-based leaf counting method of Section \ref{sec:color-based} was tested using individual perspective images, the distortion-free images, and the entire orthorectified mosaic.
By perspective images we mean the original, non-rectified individual images as they are taken from the camera.
The distortion-free images are the perspective images whose lens distortion has been removed.
Figure \ref{fig:non_rect_result_color} shows the leaf segmentation of a perspective image taken on July 15, 2015.
The local density of leaf pixels is also shown as a heat map.
The value of a given pixel in the heat map is the number of leaf pixels in the neighborhood of that pixel.
The size of the neighborhood was set to $41\times 41$ by manually selecting one of the plants.
Figure \ref{fig:result_dist_free} shows the leaf segmentation of a distortion-free image taken on July 15, 2015.
Figure \ref{fig:rect_result_color} shows the leaf segmentation of an  orthorectified mosaic.
The perspective images used to generate the mosaic were taken on June 24, 2015.
The thresholds we used for these images were experimentally
chosen as $\tau_1=30$, $\tau_2=79$, $\tau_3=30$, and $\tau_4=163$.
These threshold are HSV values in the range $[0, 0, 0]$ to
$[180, 255, 255]$ for 8 bits images.
These values may greatly change in different growth stages, and lighting conditions.

\begingroup
\makeatletter
\renewcommand{\p@subfigure}{}
\begin{figure}[h]
  \centering
  \subfigure[]
  {
          \includegraphics[width=2.6cm]{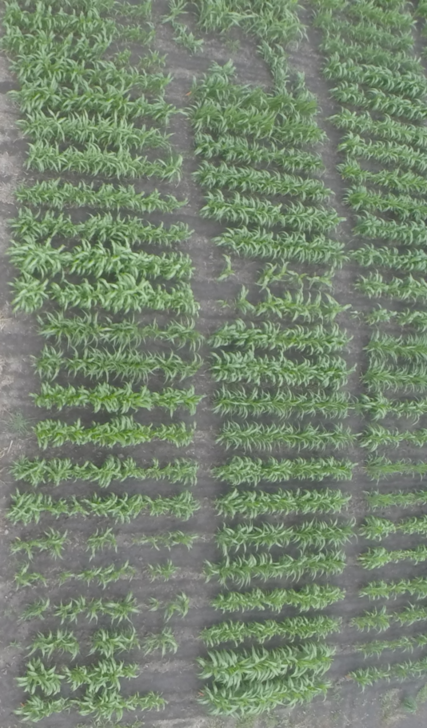}
          \label{fig:non_rect_result_color_orig}
  }
  \hspace{-0.2cm}
  \subfigure[]
  {
          \includegraphics[width=2.6cm]{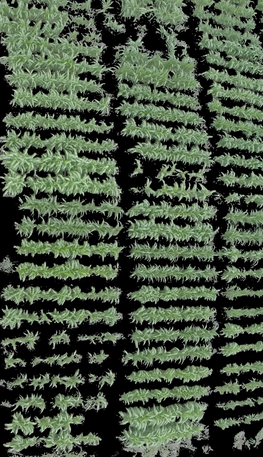}
          \label{fig:non_rect_result_color_orig_segm}
  }
  \hspace{-0.2cm}
  \subfigure[]
  {
          \includegraphics[width=2.6cm]{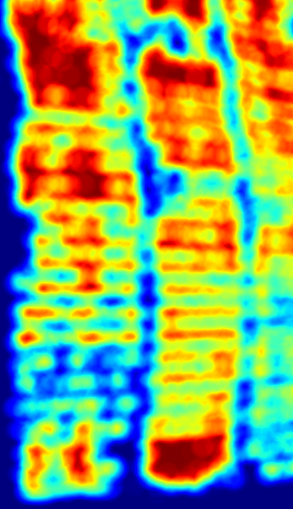}
          \label{fig:non_rect_result_color_orig_heat_map}
  }
  \hspace{-0.2cm}
  \caption{
    \ref{fig:non_rect_result_color_orig} Region of a perspective image. 
    \ref{fig:non_rect_result_color_orig_segm} Leaf segmentation using the color-based method.
    \ref{fig:non_rect_result_color_orig_heat_map} Heat map of local leaf density.
  }
  \label{fig:non_rect_result_color}
\end{figure}
\endgroup

\begingroup
\makeatletter
\renewcommand{\p@subfigure}{}
\begin{figure}[h]
  \centering
  \subfigure[]
  {
          \includegraphics[width=4cm]{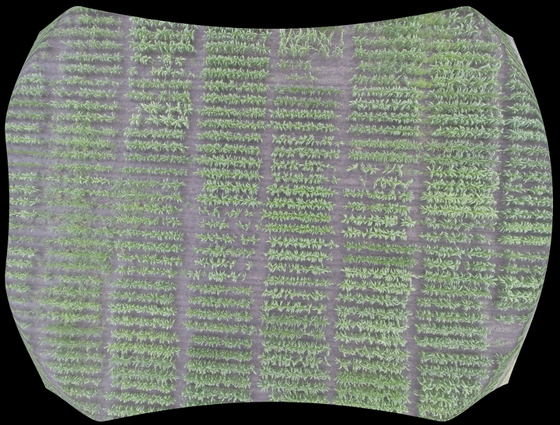}
          \label{fig:result_dist_free1}
  }
  \hspace{-0.2cm}
  \subfigure[]
  {
          \includegraphics[width=4cm]{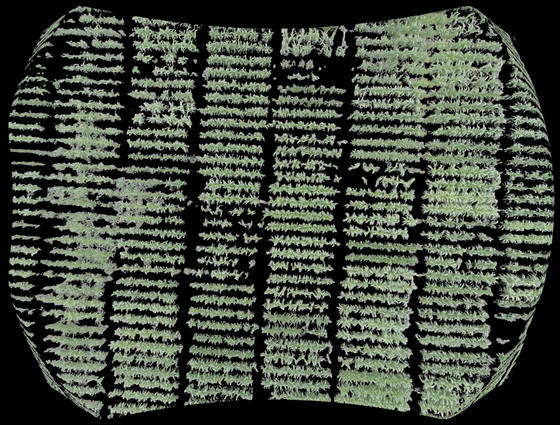}
          \label{fig:result_dist_free2}
  }
  \hspace{-0.2cm}
  \subfigure[]
  {
          \includegraphics[width=4cm]{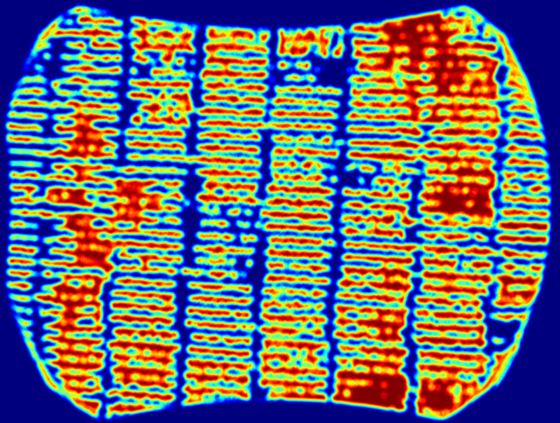}
          \label{fig:result_dist_free3}
  }
  \hspace{-0.2cm}
  \caption{
    \ref{fig:result_dist_free1} Distortion-free image.
    \ref{fig:result_dist_free2} Leaf segmentation using the color-based method.
    \ref{fig:result_dist_free3} Heat map of local leaf density.
  }
  \label{fig:result_dist_free}
\end{figure}
\endgroup


\begingroup
\makeatletter
\renewcommand{\p@subfigure}{}
\begin{figure}[h]
  \centering
  \subfigure[]
  {
          \includegraphics[width=8.4cm]{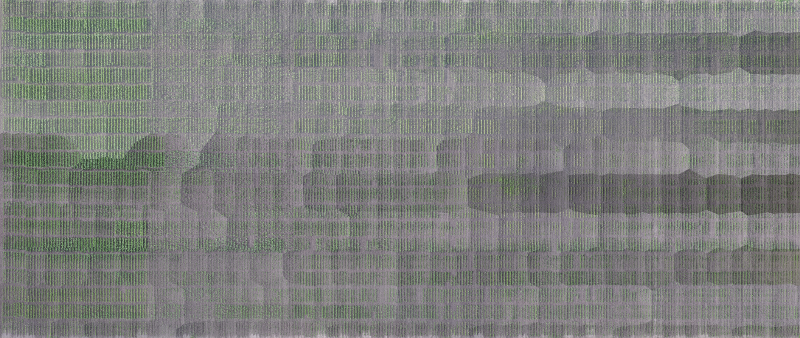}
          \label{fig:result_plot_20150715}
  }
  \hspace{-0.2cm}
  \subfigure[]
  {
          \includegraphics[width=8.4cm]{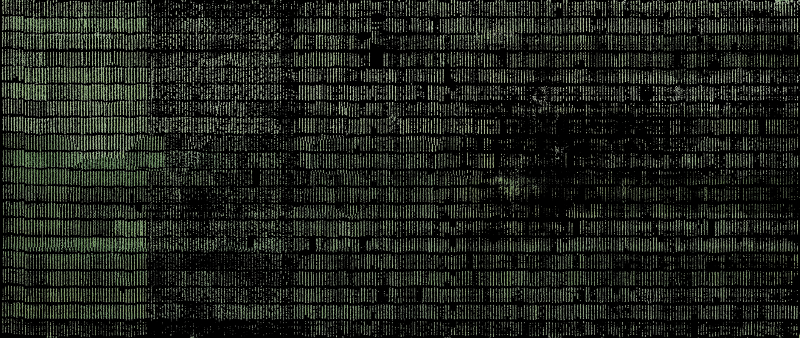}
          \label{fig:result_plot_combine}
  }
  \hspace{-0.2cm}
  \subfigure[]
  {
          \includegraphics[width=8.4cm]{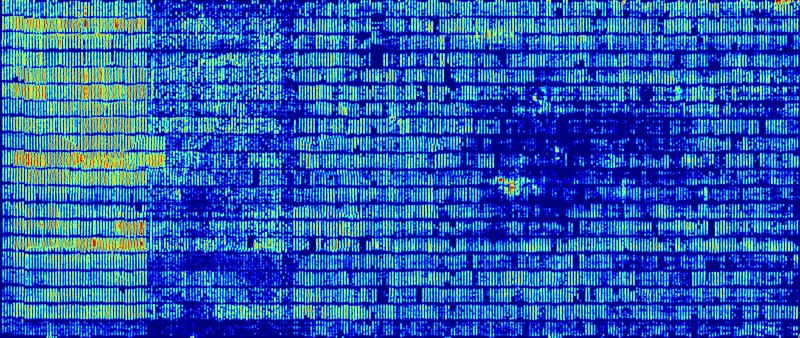}
          \label{fig:result_plot_heat_map}
  }
  \hspace{-0.2cm}
  \caption{
    \ref{fig:result_plot_20150715} Orthorectified mosaic generated by the procedure explained in Section \ref{sec:orthomethod}.
    \ref{fig:result_plot_combine} Segmented leaves.
    \ref{fig:result_plot_heat_map} Heat map of local leaf density.
  }
  \label{fig:rect_result_color}
\end{figure}
\endgroup

The leaf segmentation (Section \ref{sec:leaf_segmentation}) was used was with the orthophoto shown in Figure \ref{fig:fang_field}.
A total of 52 Sorghum rows such as the one shown in Figure \ref{fig:result_leaf_segm} were analyzed.
From the segmentation of each leaf, the leaf count was obtained.
The thresholds we used for these images were experimentally
chosen as $T_a=\frac{\pi}{5}$, $T_b=\frac{\pi}{8}$, and $T_c=\frac{\pi}{6}$.

\begingroup
\makeatletter
\renewcommand{\p@subfigure}{}
\begin{figure}[th]
  \centering
  \subfigure[]
  {
          \includegraphics[width=8.4cm]{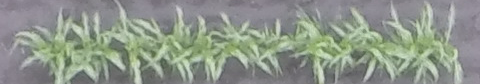}
          \label{fig:result_leaf_segm1}
  }
  \hspace{-0.2cm}
  \subfigure[]
  {
          \includegraphics[width=8.4cm]{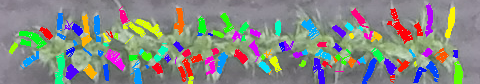}
          \label{fig:result_leaf_segm2}
  }
  \hspace{-0.2cm}
  \caption{
    \ref{fig:result_leaf_segm1} Region of a generated orthomosaic.
    \ref{fig:result_leaf_segm2} Leaf segmentation according to Section \ref{sec:leaf_segmentation}.
  }
  \label{fig:result_leaf_segm}
\end{figure}
\endgroup


Locations of plants in the orthorectified mosaic of Figure \ref{fig:fang_field} were estimated using the method proposed in Section \ref{sec:plant_location}.
Processing the entire field would take an enormous amount of time.
This is because all $N$ pixel coordinates in $Z$ are involved into the evaluation of the cost function of Equation \ref{fig:cost_function} for a single candidate $x_p$ value.
In order to reduce the computational complexity, the mosaic is cropped into regions, and each region is analyzed independently.
In Figure \ref{fig:result_location}, the cost function at one iteration is shown.
At this iteration, all plant positions are fixed except for one.
Note that the coordinate that minimizes the cost function of the free plant is very close to the true position.

\begingroup
\makeatletter
\renewcommand{\p@subfigure}{}
\begin{figure}[h!]
\centering
\subfigure[]
{
        \includegraphics[width=1.15cm]{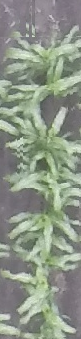}
        \label{fig:result_location_20150715}
}
\hspace{-0.2cm}
\subfigure[]
{
        \includegraphics[width=1.15cm]{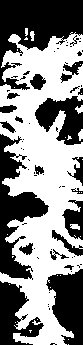}
        \label{fig:result_location_input}
}
\hspace{-0.2cm}
\subfigure[]
{
        \includegraphics[width=1.15cm]{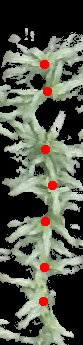}
        \label{fig:result_location_wanted}
}
\hspace{-0.2cm}
\subfigure[]
{
        \includegraphics[width=1.15cm]{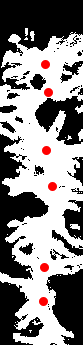}
        \label{fig:result_location_fixed_plants}
}
\hspace{-0.2cm}
\subfigure[]
{
        \includegraphics[width=1.15cm]{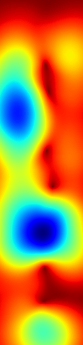}
        \label{fig:result_location_cost_map_linear}
}
\hspace{-0.2cm}
\subfigure[]
{
        \includegraphics[width=1.15cm]{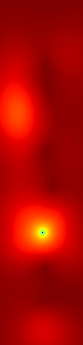}
        \label{fig:result_location_cost_map_gamma}
}
\hspace{0cm}
\caption{
  \ref{fig:result_location_20150715} Image showing 7 sorghum plants.
  \ref{fig:result_location_input} Mask $Y$, result of the color segmentation of Section \ref{sec:color-based}.
  \ref{fig:result_location_wanted} Red dots are groundtruthed plant locations.
  \ref{fig:result_location_fixed_plants} Red dots are the fixed plants at this iteration. We are estimating the location of the missing plant.
  \ref{fig:result_location_cost_map_linear} Cost function of Equation \ref{eq:k-means-cost-function} (no prior term). The minimum is selected as the location for this plant. Note that the function is non-convex.
  \ref{fig:result_location_cost_map_gamma} Cost function in gamma scale to highlight the global minimum.
}
\label{fig:result_location}
\end{figure}
\endgroup


\section{Conclusion and future work}
This paper has shown the feasibility of using consumer-grade digital cameras  onboard low-cost UAVs for the estimation of plant phenotypic traits.
Current and future work will be focusing on the use of the additional sensors such as position and orientation systems as well as LiDAR. 
Future  work will also include image-based plant height determination and the use of  leaf angle information for the estimation of the plant center location.


\section{Acknowledgments}
The information, data, or work presented herein was funded in part by the Advanced Research Projects Agency-Energy (ARPA-E), U.S. Department of Energy, under Award Number DE-AR0000593. 
The views and opinions of the authors expressed herein do not necessarily state or reflect those of the United States Government or any agency thereof.
Address all correspondence to Edward J. Delp \\(ace@ecn.purdue.edu).


\bibliographystyle{ieeetr}
\bibliography{ref}  

\end{document}